\documentclass{article}

\usepackage[english]{babel}
\usepackage{amsmath,bm,bbm,algorithmic,amssymb,hyperref,graphicx}
 
\PassOptionsToPackage{numbers, compress}{natbib}
\usepackage[final]{nips_2016}

\newcommand{\I}{\mathbbm{1}}
\newcommand{\E}{\mathbb{E}}
\newcommand{\Var}{\mathbbm{Var}}

\title{Truncation-free Hybrid Inference for DPMM}

\author{
  Arnim Bleier\\
  Department of Computational Social Science\\
  Leibniz Institute for the Social Sciences\\
  Cologne, 50667 - Germany\\
  \texttt{arnim.bleier@gesis.org}
}

\begin{document} 

\maketitle

\begin{abstract}
Dirichlet process mixture models (DPMM) are a cornerstone of Bayesian non-parametrics. While these models free from choosing the  number of components a-priori, computationally attractive variational inference often reintroduces the need to do so, via a truncation on the variational distribution. In this paper we present a truncation-free hybrid inference for DPMM, combining the advantages of sampling-based MCMC and variational methods. The proposed hybridization enables more efficient variational updates, while increasing model complexity only if needed. We evaluate the properties of the hybrid updates and their empirical performance in single- as well as mixed-membership models. Our method is easy to implement and performs favorably compared to existing schemas.
\end{abstract}

\section{Background}

To begin with, consider a model for data $\mathcal{X} = \{x_1, x_2, ..., x_N\}$ that is assumed to be generated by a mixture of simpler component models $F(\phi_k)$. Following the single-membership assumption of each data point $x_i \in \mathcal{X}$ being explained by a single component $\phi_{z_i}$ we have
\begin{align}
	\label{eq:storryline_dmm_infinite}
	\bm{\theta} &\sim Dir(\tfrac{\bm\alpha}{\bm K}) \notag \\
	\phi_k &\sim H(\beta) && \forall k \, \in [1,K] \notag \\
	z_i &\sim Cat(\bm{\theta}) && \forall i \, \in [1,N] \notag \\
	x_i &\sim F(\phi_{z_i}) && \forall i \, \in [1,N] \mbox{ ,}
\end{align}
and arrive at the infinite-dimensional DPMM for $K\to\infty$. While collapsed Gibbs sampling (CGS) is commonly used to explore the unbounded latent space of this model, consider as an alternative the collapsed variational distribution
\begin{align}
	\label{eq:variational_dist_dir}
	q(\mathbf{z}) = \prod_{i=1}^N Cat(z_i \mid \gamma_{i1},...,\gamma_{iK+1}) \mbox{ ,}
\end{align}
similar to partially collapsed approximations \citep{kurihara2007}, however with $\bm{\theta}$ and $\bm{\phi}$ integrated out. The updates to optimize the variational parameter $\bm{\gamma}_i$ with regards to the true distribution $p$ are then for each observation $i \in \{1, ..., N\}$
\begin{align}
	\label{eq:dp:gamma_updates_dir}
	\gamma_{ik} \propto 
	exp(\,\E[\,log \, p(z_i = k \mid \bm{z}_{\neg{i}})] + \E[log \, p(x_i \mid \bm{x}_{z = k}^{\neg i}, \beta)]) \mbox{ .}
\end{align}
While exact computations are hard, second-order Tailor expansions \citep{kurihara2007} or the use of zero-order information
have been suggested for estimation \cite{asuncion2009}. Using only zero-order information the optimal settings of $\bm{\gamma}$ are
\begin{align}
	\label{eq:dpmm_cvb0}
	\gamma_{ik} &\propto 
	\begin{cases} 
		  \dfrac{n_k^{\neg i}}{n_.^{\neg i} + \alpha} \; p(x_i \mid \bm{x}_{z = k}^{\neg i}, \beta) & \text{if }  k \leq K \\[1em]
		  \dfrac{\alpha}{n_.^{\neg i} + \alpha} \; p(x_i \mid \beta) & \text{if } k = K + 1 \mbox{ ,}
	\end{cases}
\end{align}
where we overload the notation $n_k^{\neg i}=\sum_{j=1}^{N\neg i}\gamma_{jk}$, as compared to the standard CGS, with the expected number of data points explained by the $k^{th}$ component. Besides the simplifying assumptions made, these updates would place probability mass on each component $\gamma_{ik} \neq 0 \; \forall k \in \{1,...,K+1\}$ and introduce a new component in each step of the inference. A common way to address this problem is the use of fixed finite-dimensional variational approximations for DPMM \citep{kurihara2007}. Alternatively, Lin \citep{lin2013} as well as Wang et al. \citep{wang2012}, amongst others, discuss methods for growing the truncation as part of the inference. Lin \citep{lin2013} introduces an additional parameter controlling the growth of the truncation. Wang et al. \citep{wang2012} estimate parameters for \emph{locally collapsed variational inference} from traditional samples, losing valuable information in the updates.

We extend the idea of estimating variational parameters from samples with a method to construct the samples more efficiently. In the remainder of this paper, we start with the construction of the proposed truncation-free hybrid updates. After that we study the properties of these updates. We conclude with a short evaluation as well as a discussion of the current limitations and directions for future work.

\section{Construction of Hybrid Updates}

\begin{figure}
	\centering
	\includegraphics[width=0.52\textwidth]{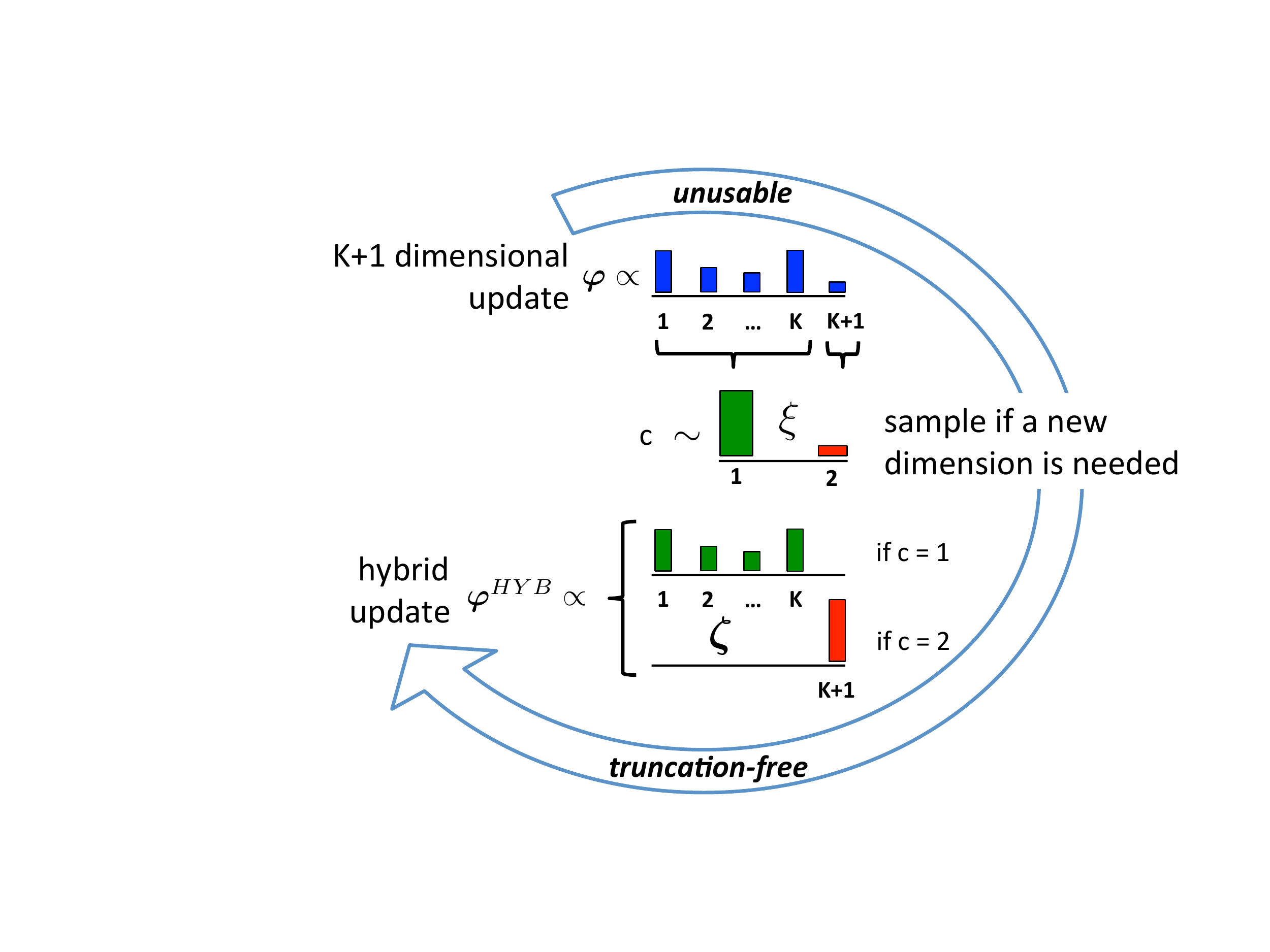}
	\caption{Illustration of hybrid updates}
	\label{fig:construction}
\end{figure}

Our goal is to allow for a truncation-free variational posterior inference that keeps as much information as possible of the updates, while still being able to explore the unbounded latent space in a fashion similar to the Gibbs sampler. To reach that goal, we suggest to replace the unusable $K+1$ dimensional update of the variational parameter $\bm{\gamma}_{i}$ with a hybrid update. The suggested update has either the form of a $K$ dimensional truncated variational parameter or a $K+1$ dimensional parameter instantiating a new component. Our hybridization depends only on local information and we will use the abbreviation $\varphi_k = \gamma_{ik}$ to refer, in this section, to the $k^{th}$ component in the $K+1$ dimensional probability vector in Equation~\ref{eq:dpmm_cvb0}. 

Let $\bm{\xi}$ be the two-dimensional parameter of a categorical distribution with the first dimension
\begin{align}
	\label{eq:varsigma_1}
	\xi_1 \propto \sum_{k=1}^K\varphi_{k}
\end{align}
being proportional to the sum of the explanatory power of the first $K$ components, and the second dimension
\begin{align}
	\label{eq:varsigma_2}
	\xi_2 \propto \varphi_{K+1}
\end{align}
being proportional to what is explained by the yet uninstantiated components. Furthermore, let $\bm{\zeta}_c$, with $c \in \{1,2\}$, be probability vectors representing, respectively, a truncated variational distribution and a Gibbs sample instantiating a new dimension in vector notation
\begin{align}
	\label{eq:hybrid_1}
	\zeta_{1k} &\propto 
	\begin{cases} 
		\varphi_{k} & \text{if }  k \leq K \\[1em]
		0 & \text{if } k = K + 1
	\end{cases} \mbox{ ,} &&&
	\zeta_{2k} &\propto 
	\begin{cases} 
		0 & \text{if }  k \leq K \\[1em]
		1 & \text{if } k = K + 1 \mbox{ .}
	\end{cases} 
\end{align}
With this setup, we then sample a variable
\begin{align}
	\label{eq:hybrid_c}
	c \sim  Cat(\bm{\xi})
\end{align} 
from $\bm{\xi}$ to indicate whether the truncated variational update or the probability vector instantiating a new component is selected
\begin{align}
	\label{eq:hybrid_11}
	\varphi_{k}^{HYB} \propto \zeta_{ck}\mbox{  .}
\end{align}
The probability vector $\bm{\varphi}^{HYB}$ is our hybrid update. The hybrid update replaces the original unusable $K+1$ dimensional variational update $\bm{\gamma}_{i} \leftarrow \bm{\varphi}^{HYB}$, using most of the time the efficient truncated $K$ dimensional variational distribution without introducing a new dimension in the update step, while introducing a new $K+1^{th}$ component only if needed, similar to a Gibbs sampler. For a graphical illustration of the construction of $\bm{\varphi}^{HYB}$, see Figure~\ref{fig:construction}.

\section{Properties of Hybrid Updates} 

\begin{figure}
	\centering
	\includegraphics[width=0.45\textwidth]{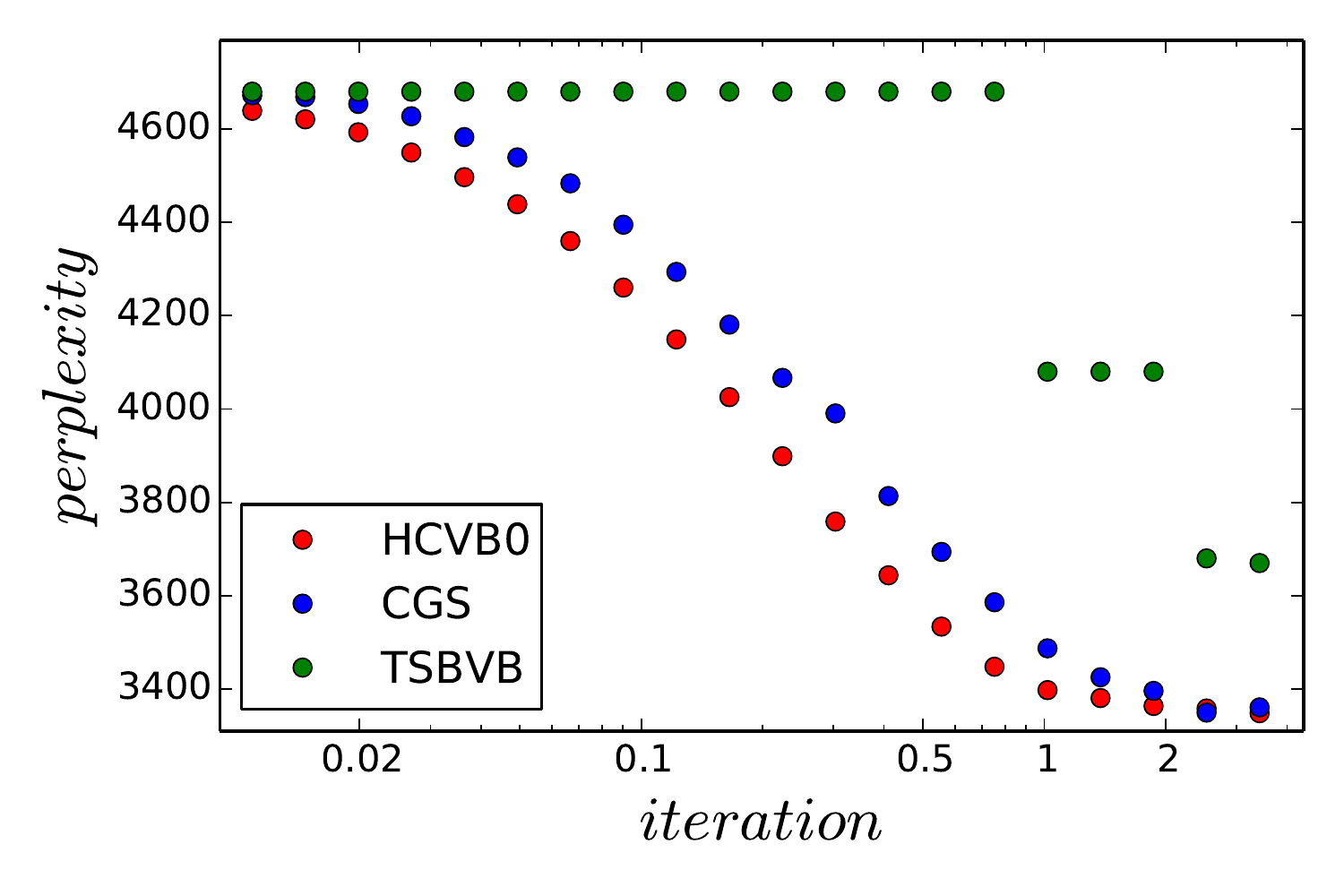}
	\caption{Predictive performance in singe-membership models for the Associated Press data set.}
	\label{fig:ap_ppx}
\end{figure}

This construction of the hybrid updates has a number of favorable properties. By definition of the parameter $\bm{\xi}$ (Equations~\ref{eq:varsigma_1},~\ref{eq:varsigma_2}), the event of sampling the second category $c=2$ has the same expectation and variance as sampling a new component in the Gibbs sampler:
\begin{align}
	\label{eq:prop_hyb_2}
	\E[\I[c = 2]] &= \E[\I[z_{i} = K+1]] \mbox{  } \notag \\\
	\Var[\I[c = 2]] &= \Var[\I[z_{i} = K+1]]\mbox{  .} 
\end{align}
Together with Equation~\ref{eq:hybrid_1} and \ref{eq:hybrid_11}, we see that this carries over to the hybrid update itself 
\begin{align}
	\label{eq:prop_hyb_3}
	\E[\I[\varphi_{K+1}^{HYB} = 1]] &= \E[\I[z_{i} = K+1]] \mbox{  } \notag \\\
	\Var[\I[\varphi_{K+1}^{HYB} = 1]] &= \Var[\I[z_{i} = K+1]]\mbox{  ,} 
\end{align}
making it possible to introduce new components like in a Gibbs sampler. Even more, the preservation of expectation is not limited to the $K+1^{th}$ dimension itself, but our hybrid updates $\bm{\varphi}^{HYB}$ preserve the expectation, with regards to $\varphi_{k}$, over all $K+1$ dimensions
\begin{align}
	\label{eq:prop_hyb_1}
	\E[\varphi_{k}^{HYB}] = \E[\varphi_{k}] \;\; \forall k \in [1,..,K+1] \mbox{ .}
\end{align}
Note that this is, with the exception of \textit{locally collapsed variational inference} \citep{wang2012}, generally not the case for variational updates in non-parametric models. 

Moreover, the sum of the explanatory power of the existing $K$ dimensions will exceed the explanatory power of introducing a new dimension ${\E[\xi_1] > \E[\xi_2]}$ for most data points, supporting the use of the more informative variational distribution $\bm{\zeta}_{1}$ in most of the updates. Finally, the computations necessary for the hybrid update $\bm{\varphi}^{HYB}$ are easy to implement and readily available by almost no additional computational costs from the normalization terms of $\bm{\varphi}$.

\newpage\section{Experiments and Discussion}

This section concludes our paper with an early empirical evaluation of the proposed hybrid updates. For the evaluation we used two text data sets: (1) The \textit{Associated Press} corpus consisting of $2, 250$ documents, where we used a vocabulary of $10,932$ distinct terms occurring over a total of $398k$ tokens. (2) The larger \textit{New York Times} corpus consisting of $1,8$ million articles, from which we extracted $153$ million tokens using a vocabulary of $77,928$ distinct terms.

The \textit{Associated Press} corpus was used for evaluating the proposed updates in the single-membership model together with a Dirichlet-Multinomial data model for the documents. In the experiments, we held out 20\% of the documents as a test set $\mathcal{X}^{test}$ and batch-trained on the remaining $\mathcal{X}^{train}$ documents. Next, we split each test document $\bm{x}_i \in \mathcal{X}^{test}$ in two parts $\bm{x}_i = (\bm{x}_i^{a}, \bm{x}_i^{b})$, $\bm{x}_i^{a}$ consisting of 70\% of the document for estimating the indicator variable and computed the perplexity of the remaining 30\% $\bm{x}_i^{b}$. We then compared the perplexity versus the number of iterations. Figure \ref{fig:ap_ppx} displays the performance for hybrid updates in the zero-order collapsed variational setting (HCVB0), CGS and truncated (T = 40) stick-breaking mean field variational Bayes (TSBVB) \citep{kurihara2007}. 

\begin{figure}
	\centering
	\begin{center}
		\begin{tabular}{cc}
			\includegraphics[width=0.47\textwidth]{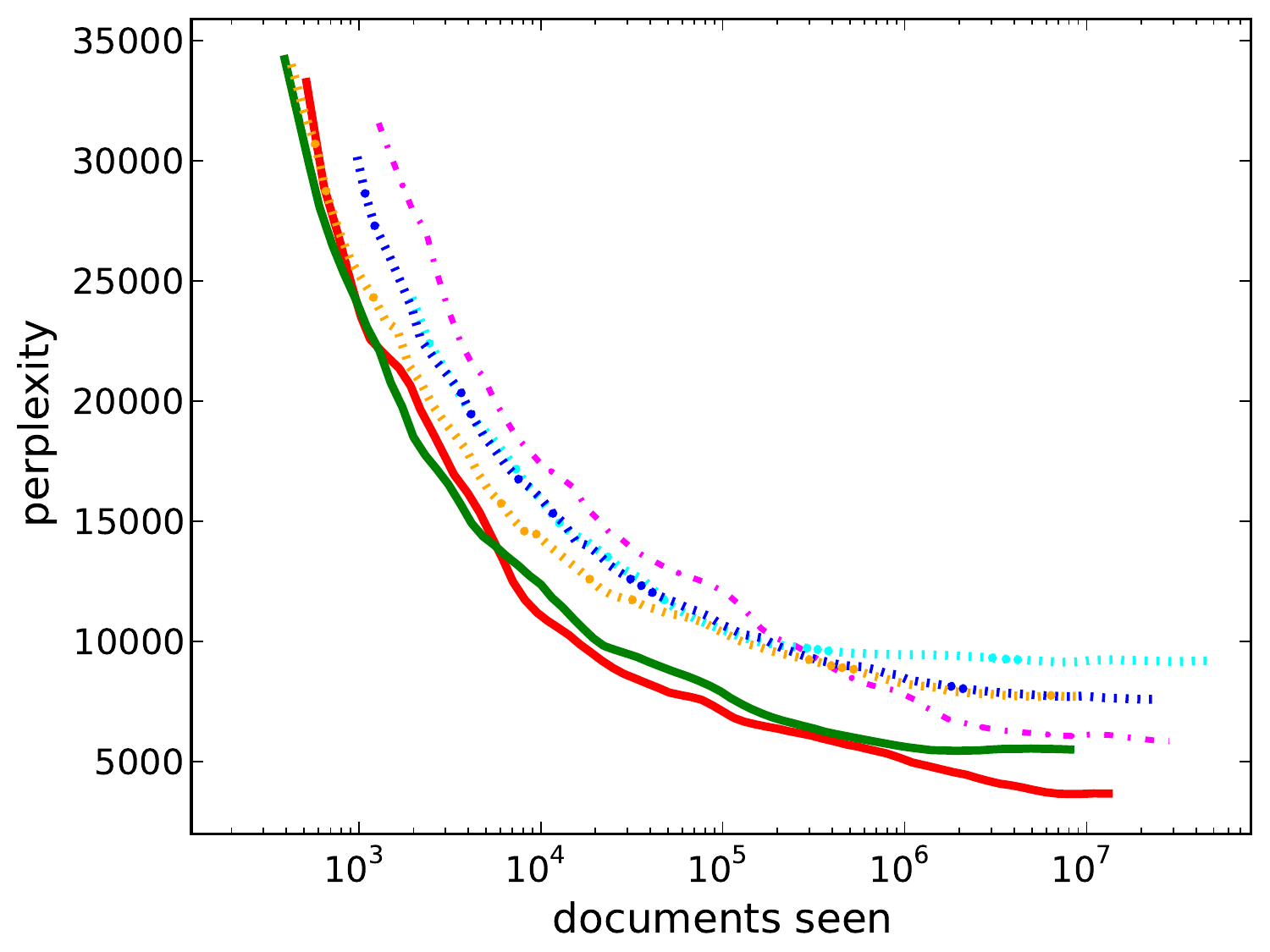} &
			\includegraphics[width=0.47\textwidth]{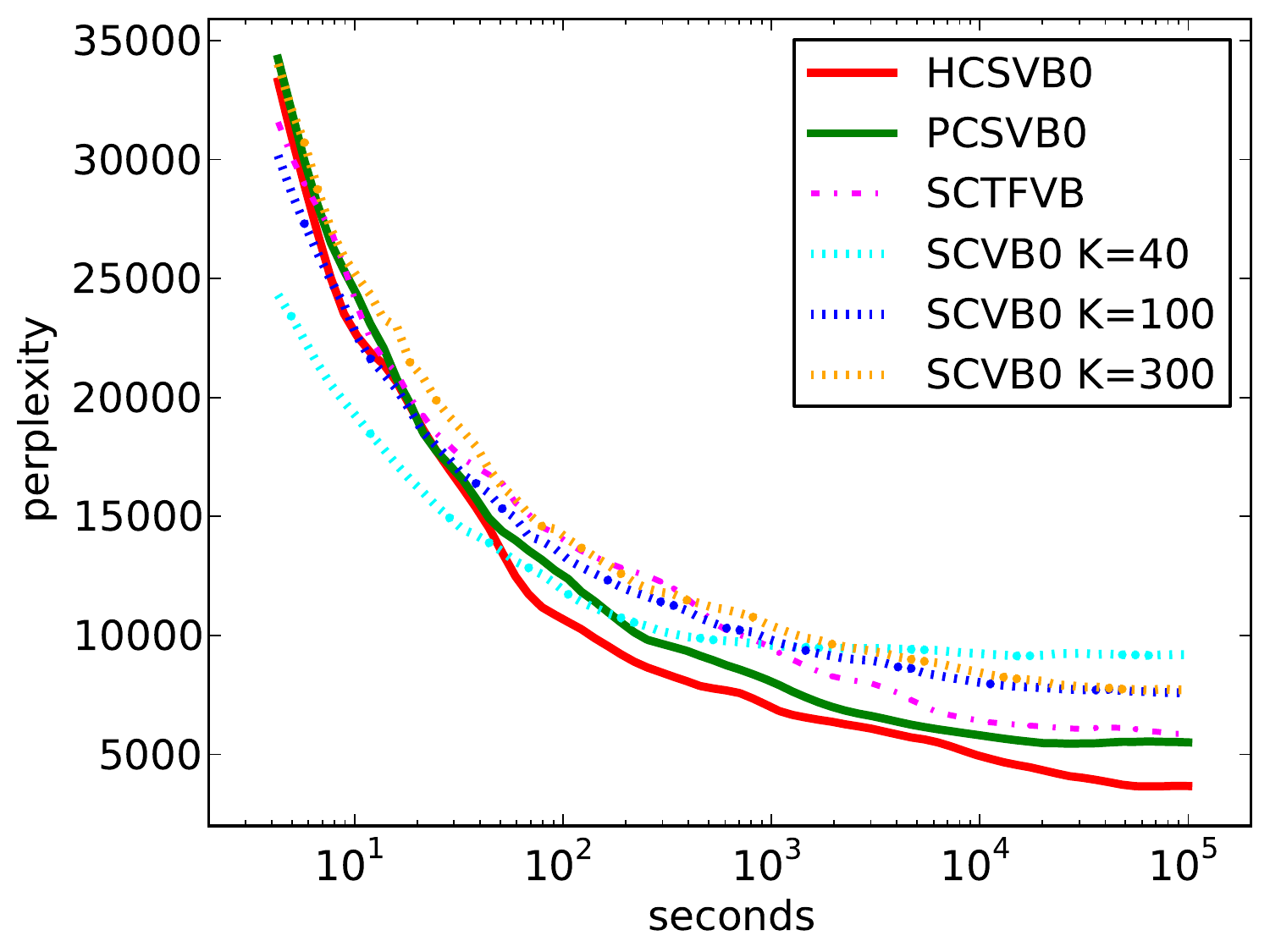} 
		\end{tabular}
	\end{center}
	\caption{Predictive performance in HDP-LDA models for the New York Times data set.}
	\label{fig:hybrid_ppx}
\end{figure}

The \textit{New York Times} corpus was used for evaluating the hybrid updates in mixed-membership HDP-LDA models, with test-train splits similar to above. However, for this larger dataset we resorted to collapsed stochastic inference using minibatches of 60 documents. We compared our method to the truncation-free \emph{locally collapsed variational inference} (SCTFVB) \citep{wang2012} and the finite dimensional stochastic collapsed variational Bayesian inference for LDA (SCVB0) \cite{foulds2013} using 40, 100 and 300 topics. We employed our hybrid updates in a setting similar to SCTFVB, however using a lower-bound approximation for the estimation of the stick-breaking weights \cite{sato2012, bl2013b}. We used the same parameterizations for the update schedules in all inference schemas. Figure \ref{fig:hybrid_ppx} displays the results, as a function of the number of documents processed (left) and wall-clock time in seconds (right), for the same runs. In the figure, HCSVB0 denotes the results with our hybrid updates. PCSVB0 denotes a finite dimensional variational approximation otherwise identical to HCSVB0, but truncated at 300 topics.


In this work, we desired to use the advantages of MCMC schemas and variational schemas combined in a single inference schema for Bayesian non-parametric models. We answered to this demand by presenting a novel type of hybridization that efficiently uses the full variational distribution while sampling for the introduction of new components.
The proposed method is easy to implement and measurably improves the predictive performance over state of the art methods for single- as well as mixed-membership models at little additional computational cost. The current limitations of the presented work are two-fold. While we have established some favorable properties of the updates and found predictive performance improvements, we rely on approximations and have only limited theoretical arguments legitimizing our approach. The other limitation of this work is its scope. Next to a more thorough experimental evaluation and further formalization, an adaption of the hybrid updates to Wang et al. \cite{wang2011}'s Chinese restaurant process based variational inference for the HDP could potentially be a promising direction for future work.

\clearpage

\bibliography{bibliography}

\end{document}